\documentclass{article}
\usepackage{ijcai20}

\usepackage{times}
\usepackage{soul}
\usepackage{url}
\usepackage[utf8]{inputenc}
\usepackage[small]{caption}
\usepackage{graphicx}
\usepackage{multirow}
\usepackage{subfigure}
\usepackage{amsmath}
\usepackage{amsthm}
\usepackage{booktabs}
\usepackage{algorithm}
\usepackage{algorithmic}
\usepackage{amsfonts}
\usepackage{color}
\newcommand{\newcite}[1]{\citeauthor{#1}~\shortcite{#1}}

\title{Dialogue State Induction Using Neural Latent Variable Models}

\author{
	Qingkai Min$^{1,2}$
	\and
	Libo Qin$^3$\and
	Zhiyang Teng$^{1,2}$\and
	Xiao Liu$^{4}$\And
	Yue Zhang$^{1,2}$
	\affiliations
	$^1$School of Engineering, Westlake University\\
	$^2$Institute of Advanced Technology, Westlake Institute for Advanced Study\\
	$^3$Research Center for Social Computing and Information Retrieval, Harbin Institute of Technology\\
	$^4$School of Computer Science and Technology, Beijing Institute of Technology
	\emails
	\{minqingkai,tengzhiyang,zhangyue\}@westlake.edu.cn,
	lbqin@ir.hit.edu.cn,
	xiaoliu@bit.edu.cn
}

\begin{document}

\maketitle

\begin{abstract}
Dialogue state modules are a useful component in a task-oriented dialogue system. Traditional methods find dialogue states by manually labeling training corpora, upon which neural models are trained. However, the labeling process can be costly, slow, error-prone, and more importantly, cannot cover the vast range of domains in real-world dialogues for customer service. We propose the task of dialogue state induction, building two neural latent variable models that mine dialogue states automatically from unlabeled customer service dialogue records. Results show that the models can effectively find meaningful dialogue states. In addition, equipped with induced dialogue states, a state-of-the-art dialogue system gives better performance compared with not using a dialogue state module.
\end{abstract}

\section{Introduction}
\begin{figure}[!t]
	\centering
	\includegraphics[width=0.48\textwidth]{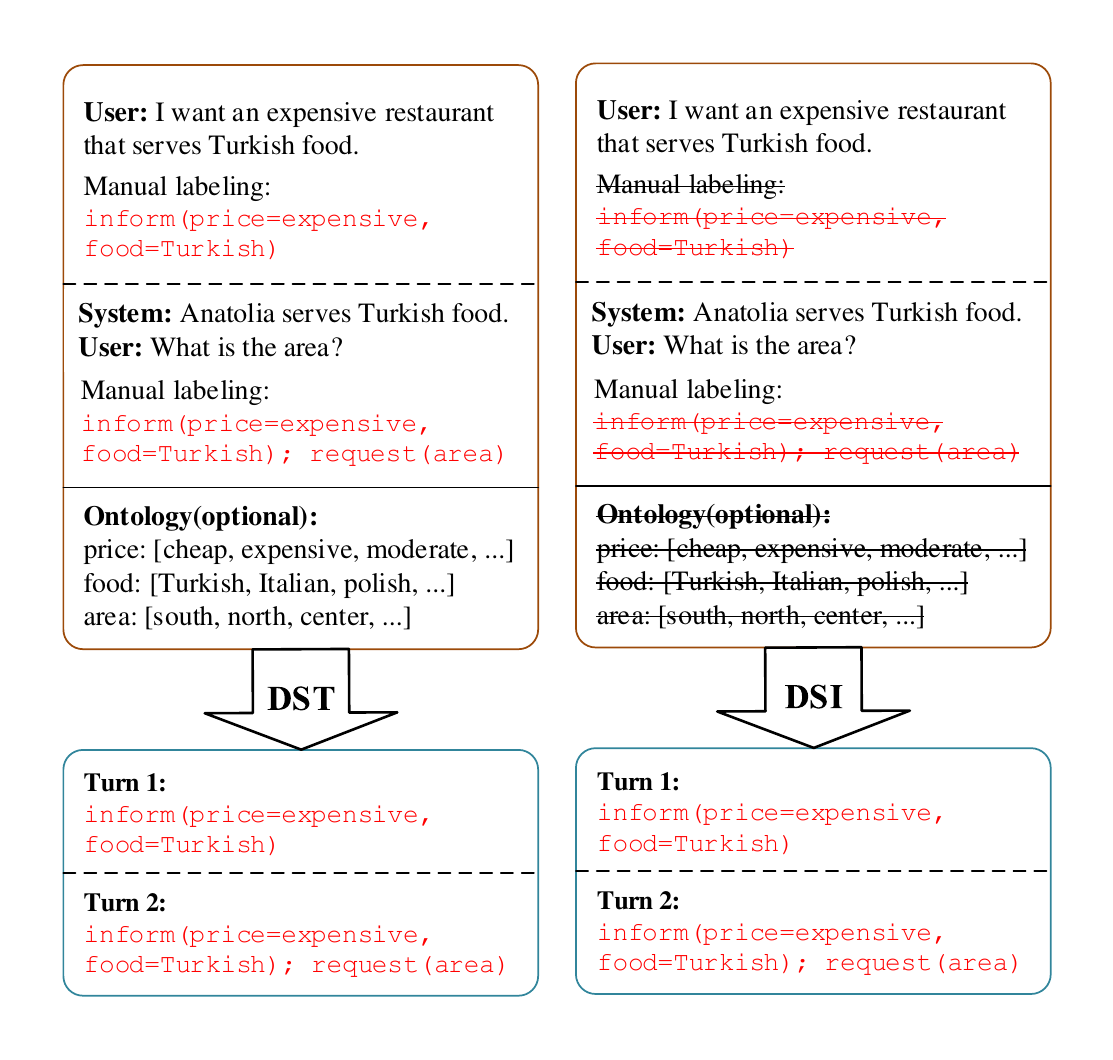}
	\caption{Comparison between DSI and traditional DST. The strikethrough font is used to represent the resources not needed by DSI. The dialogue state is accumulated as the dialogue proceeds. Turns are separated by dashed lines. Dialogues and external ontology are separated by black lines.}
	\label{fig:architecture}
\end{figure}

Dialogue state modules are a central component to a task-oriented dialogue system~\cite{wen2017network,lei-etal-2018-sequicity}. Given a user utterance and existing dialogue history, a dialogue system typically extracts dialogue states, according to which a system response is generated. An example is shown in Figure~\ref{fig:architecture}, given two turns of a dialogue, the first user utterance is ``I want an expensive restaurant that serves Turkish food.'', and the dialogue states consist of the slot-value pairs \texttt{inform(price=expensive, food=Turkish)}. As the dialogue proceeds, the dialogue state is updated at each turn. After tow dialogue turns, the dialogue state becomes \texttt{inform(price=expensive, food=Turkish); request(area)}, where \textit{inform} represents the search constraints expressed by user and \textit{request} represents the search target that the user is asking for. In this example, the user intention is to reserve a restaurant. The business domain is restaurant customer service. The dialogue state represents what the user is looking for at the current turn of the conversation. 



Prior work has mostly followed a manual labeling-train-test paradigm, which begins with the design of annotation guidelines, followed by the collection and manual labeling of training corpora, before training a model.  The supervised learning task is called {\it Dialogue State Tracking} (DST)~\cite{young2010hidden}. One limitation for supervised learning is that the manual labeling process can be slow and costly given a certain domain of customer service. Available datasets are labeled on a few popular domains such as restaurant, taxi, train and hotel~\cite{williams2014dialog,budzianowski-etal-2018-multiwoz,rastogi2019towards}. However, in practice, the number of customer service domains ranges far beyond hundreds (e.g. telecommunication customer service, banking, household maintenance, police, e-commercial customer service, etc), which makes it infeasible to manually label corpora for every domain. In addition, it has been shown that the ratio of annotation errors can be as high as 30\% and even 40\% for the DST task (i.e., the MultiWOZ datasets)~\cite{eric2019multiwoz,zhang2019classify}.


To address this issue, it can be useful to automatically induce dialogue states from raw dialogues. We assume that there is a large set of dialogue records of many different domains, but without manual labeling of dialogue states. Such data are relatively easy to obtain, for example from customer service call records from different businesses. Consequently, we propose the task of {\bf dialogue state induction} (DSI), which is to automatically induce slot-value pairs from raw dialogue data and can be better used for downstream dialogue tasks such as database query, act prediction and response generation. The difference between DSI and DST is illustrated in Figure~\ref{fig:architecture}.  Similar to DST models, DSI outputs dialogue states in slot-value pairs such as \texttt{inform(price=expensive)}, where {\it price} represents a slot, and {\it expensive} represents a value. For requestable slots such as \texttt{request(area)}, \textit{request} is regarded as the slot and \textit{area} is regarded as the value. During training, DST relies on both a dialogue record and manual labeling of slot-value pairs on the dialogues. In contrast, our task does not rely on manual labeling and can generate slot-value pairs over raw dialogues automatically. 

We introduce two neural latent variable models for DSI by treating the whole state and each slot as latent variables, from which values observed in dialogue data are generated. The goal is to induce slots according to those frequently co-occurring values and the dialogue contexts. In particular, each value (i.e., phrase in raw text) is represented by using both a sparse one-hot representation and a dense contextualized embedding representation. Both models are generative probabilistic models, which generate a value by first generating a latent dialogue state vector, and then generating a slot. The difference between the two models is the modeling of service domains. We observe that different service domains may contain slots with similar contexts or values. For example, both {\it taxi} and {\it bus} domains can have the same slot {\it to\_location}. In order not to mix their structures from a large dialogue record, our second model further considers the service domain explicitly by taking the dialogue state as a {\it Mixture-of-Gaussians}. We refer to the basic model {\it DSI-base} and the advanced model {\it DSI-GM}.


Experiments over the MultiWOZ~\cite{budzianowski-etal-2018-multiwoz} and the SGD~\cite{rastogi2019towards} datasets show that both DSI models can effectively induce dialogue states compared with a random select strategy. In addition, the Gaussian mixture model gives significantly better results compared with the basic model. Finally, we apply DSI to a recently proposed dialogue system~\cite{chen-etal-2019-semantically}, by replacing the dialogue state module with our {\it DSI-GM} model. Results show that adding induced dialogue states gives significantly better results in both dialogue act prediction and response BLEU compared with a dialogue system without considering dialogue states. In particular, the BLEU score using the {\it DSI-GM} model outputs is better by 2.1\% compared with not using dialogue states, and lower by 0.8\% compared with using manual labeling dialogue states. This shows that by inducing dialogue states, improved dialogue systems can be obtained. To our knowledge, we are the first to automatically induce dialogue states in the form of slot-value pairs using a neural latent variable model. Our models can serve as baselines for further research. We release our code at \url{https://github.com/taolusi/dialogue-state-induction}.

\section{Related Work}

{\bf The role of DST and DSI in task-oriented dialogue systems}. Task-oriented dialogue systems are complex, traditionally involving a pipeline of multiple steps, including automatic speech recognition (ASR)~\cite{wen2017network}, spoken language understanding (SLU)~\cite{qin-etal-2019-stack}, dialogue state tracking (DST)~\cite{zhong2018global}, policy learning and natural language generation (NLG)~\cite{chen-etal-2019-semantically}. SLU consistes of two main sub-tasks, namely intent detection, which is to identify the user intent such as \textit{hotel booking}, and slot tagging, which is to identify relevant semantic slots in a user utterance, such as \texttt{price} and \texttt{stars}. Dialogue state tracking aims to identify user goals at every turn of the dialogue, such as \texttt{inform(price=moderate, stars=4); request(phone)}, which makes the core component in a task-oriented dialogue system. Policy learning aims to learn the system action based on the current state. Natural language generation transforms the system action into natural language.

Recently, some work on task-oriented dialogue systems takes an overall end-to-end method, by encoding the user utterance and dialogue history, and then generating a response directly using a seq2seq model variant, without explicitly maintaining dialogue states~\cite{eric2017copy,2020arXiv200411019Q}. Compared to such work, we show that automatically inducing dialogue states can improve dialogue performance, which is consistent with observations of DST research~\cite{lei-etal-2018-sequicity,wen-etal-2018-sequence}.  \newcite{wen2017network} pioneered this line of work by proposing a typical modularly connected end-to-end trainable task-oriented dialogue system directly based on text without considering the speech recognition noise and thus ignored the component of SLU to obtain the dialogue state. 

{\bf DST vs SLU}.
In a traditional pipeline, DST operates on SLU output to update the dialogue state dealing with noise from ASR and SLU. In particular, SLU can give an N-Best list of semantic representations based on the N-Best list of sentences from ASR. DST handles all these uncertainties, e.g. error propagation, to update the dialogue state. However, the recent datasets are collected based on text without taking noisy speech inputs into consideration, which has made the task of slot tagging in SLU and the task of DST rather separately investigated. The correlation between recent slot tagging work and DST work can be subtle, and there is little discussion in the literature about their fundamental differences. In practice, widely used datasets for SLU include ATIS~\cite{hemphill1990atis}, while for DST include MultiWoZ~\cite{budzianowski-etal-2018-multiwoz}. For latter datasets, there is not labeling of the SLU task. Recently, DST research has taken end-to-end methods, without considering SLU as a pre-processing step. Compared with SLU, our work is more in line with current neural DST work, with DSI being a direct alternative to DST in zero-shot training scenarios.


{\bf DSI vs robust DST}.
Existing work on DST integrates dialogue states by manually labeling more or less. Compared to these methods, our method has exactly the same setting as end-to-end DST, which is more cost-effective. Traditional supervised models for dialogue state tracking regard the task as a multi-class classification problem~\cite{mrkvsic2015multi}. Given a user utterance and existing dialogue history, a model predicts the corresponding value (or None) of each slot. However, such methods cannot predict the existence of unknown slot values. Consequently, recent work begins to investigate value generation from scratch~\cite{ren2019scalable}. Such methods reduce the decoding complexity by avoiding the enumeration of all possible slots and values. However, the models still rely on supervised data for training. Our method employs the same decoding efficiency, yet additionally does not require labeled corpora. 

DSI is also related to domain adaptations of DST in handling unknown data. Supervised methods consider multi-domain settings by parameter sharing~\cite{WuTradeDST2019}. Such methods cannot deal with unknown domains, which are dominant in practice. Some work considers zero-shot learning, transferring knowledge from known domains to unknown domains without labels~\cite{rastogi2019towards}. One constraint of such methods is that they rely on domain similarity for transfer, and therefore cannot be applied to general domains. In addition, they rely on schema-level slot descriptions for capturing domain correlation, which requires manual labeling and for which the quality is difficult to control. In contrast, our method can be directly used to induce dialogue states from arbitrary dialogue records.

\begin{figure}[!t]
    \centering 
	\includegraphics[width=0.48\textwidth]{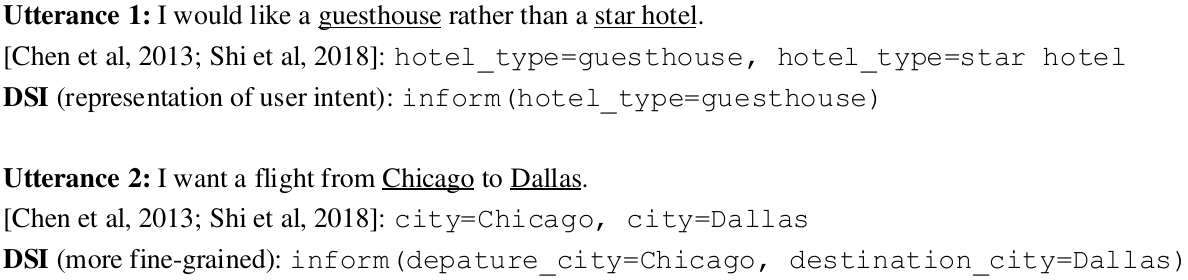}
	\caption{Comparison between DSI and previous induction methods for SLU.}
	\label{fig:DSI-SLU}
\end{figure}

{\bf Induction methods}.
The closest in spirit of our work, \newcite{chen2013unsupervised} used the FrameNet-style frame-semantic parsers to induce slots from a user utterance; \newcite{shi2018auto} proposed a framework \textit{auto-dialabel} to cluster noun words into slots. Compared with their work, our work is different in two main aspects. First, the problems that we solve are different, which can be seen in Figure~\ref{fig:DSI-SLU}. In particular, given the utterance ``I would like a guesthouse rather than a star hotel.'', the user intent is to book a hotel, the slots include \texttt{hotel\_type=guesthouse} and \texttt{hotel\_type=star hotel}, and the dialogue state is \texttt{inform(hotel\_type=guesthouse)}. Given the sentence ``I want a flight from Chicago to Dallas'', the user intent is to book a flight, the slots include \texttt{city=Chicago} and \texttt{city=Dallas}, and the dialogue state is \texttt{inform(departure\_city=Chicago, destination\_city=Dallas)}. From the two examples we can see that DSI not only reflects the user goals but also is more specific to the current dialogue state, while their methods are more general to the semantics. Our task is directly useful for subsequent policy learning and response generation tasks. Second, we consider a deep neural model with hidden variables and contextualized embeddings, which also adapts better to the multi-domain scenario. 

\section{Task Definition: Dialogue State Induction}

Given a set of customer service records without annotation (e.g. user intent and dialogue states or other manual labeling), the task is to automatically discover information that the user is looking for at each turn. We call this automatic discovery process \textit{dialogue state induction} (DSI). In particular, at each turn, the current user utterance and dialogue history can be used as input, and the output is a set of slot-value pairs, namely, the dialogue state.

\paragraph{Turn-level vs joint-level dialogue state.}
By definition, a dialogue state should reflect the user goal from the beginning of the dialogue until the current turn. We call this a joint-level dialogue state. In practice, DST research has also investigated the extraction of local user goal at each dialogue turn, which we call a turn-level state.
Our models produce dialogue state for each dialogue turn, where the input is the current user utterance and its preceding system utterance, and the output is a set of slot-value pairs. For multi-turn dialogues, the current dialogue state should reflect the whole dialogue history, as problem definition specifies. Following~\newcite{zhong2018global}, we handle this issue by simply using the union of slot-value pairs in each history dialogue turn for representing the current dialogue state. When there are multiple values for one slot, we use the latest value.

\section{Method}
We build two incrementally more complex neural latent variable models for DSI. The models induce dialogue state over a dialogue turn according to a generate process from slots to candidate values, where candidate values are represented by both one-hot vectors and contextualized embedding vectors. The two types of representations are complementary to each other, with the latter also containing features from a global context. In particular, the current user utterance and its preceding system utterance are concatenated (in their chronical order, with the latter before the former) and fed as input to a pre-trained ELMo model to obtain the contextualized word embedding vector\footnote{In practice, we use the ELMo pre-trained model with the output size of 256. We also tried BERT, whose performance is almost the same as ELMo. However, BERT is much slower and more resource-intensive for training.}. 

\subsection{Values}
Both methods are generative models that induce slot-value pairs over candidate values over a dialogue dataset. 
We follow \newcite{goel2019hyst} and extract possible values from local conversation contexts before assigning them to slots. We take a different method to this end. 
In particular, we use the model of~\newcite{cui-zhang-2019-hierarchically} to extract POS tags and the Stanford CoreNLP toolkit to extract named entities and coreferences. A set of rules are used to extract candidate values given the POS and entity patterns including filtering stopwords, repeated candidates and non-representative entity mentions. 

\begin{figure}[t]
	\setlength{\belowcaptionskip}{-0.35cm}
    \centering
    \subfigure[\textit{DSI-base}]{
        \centering
        \includegraphics[width=0.2\textwidth]{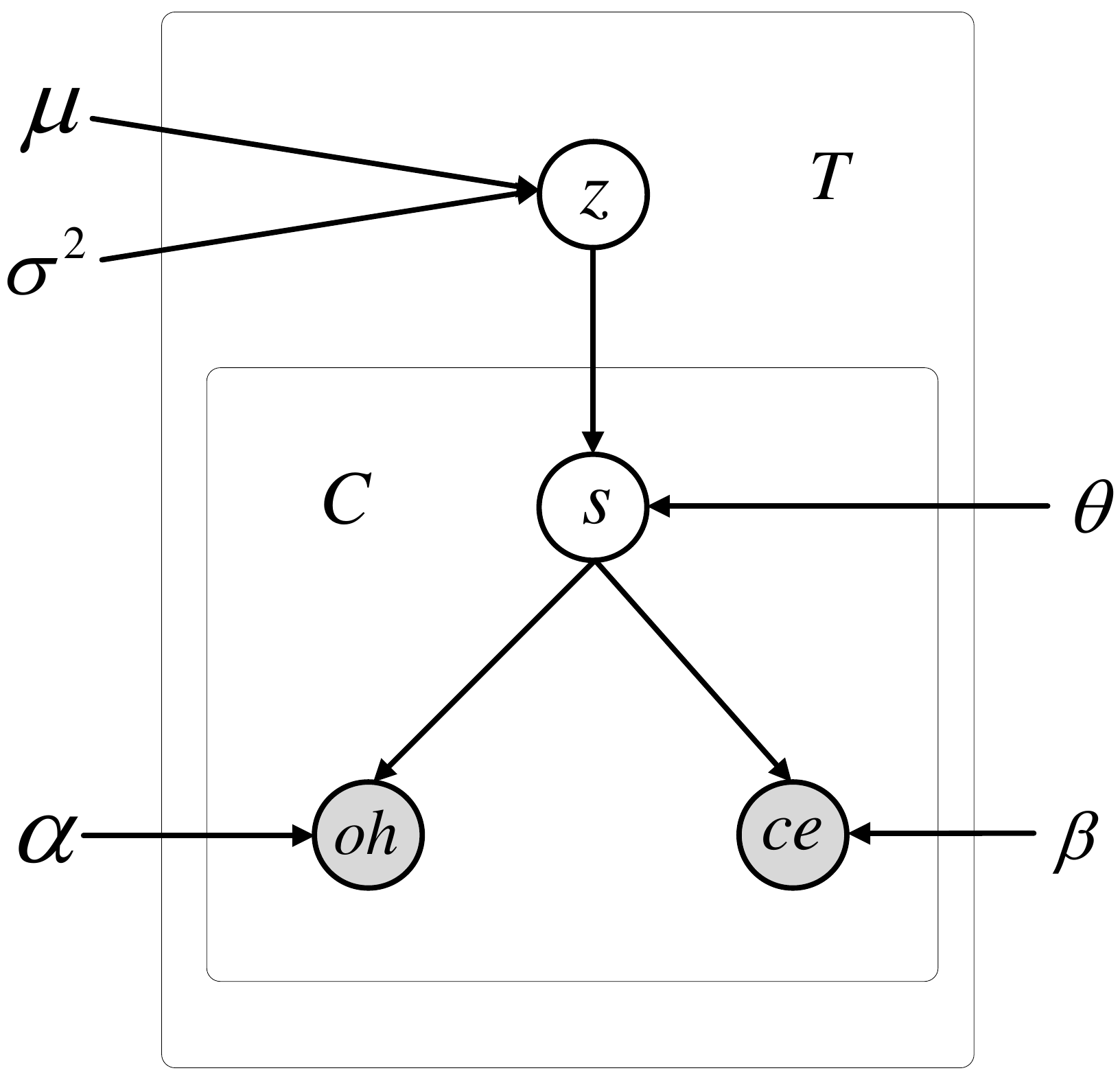}
        \label{fig:model_vae}
    }
    \subfigure[\textit{DSI-GM}]{
        \centering
        \includegraphics[width=0.22\textwidth]{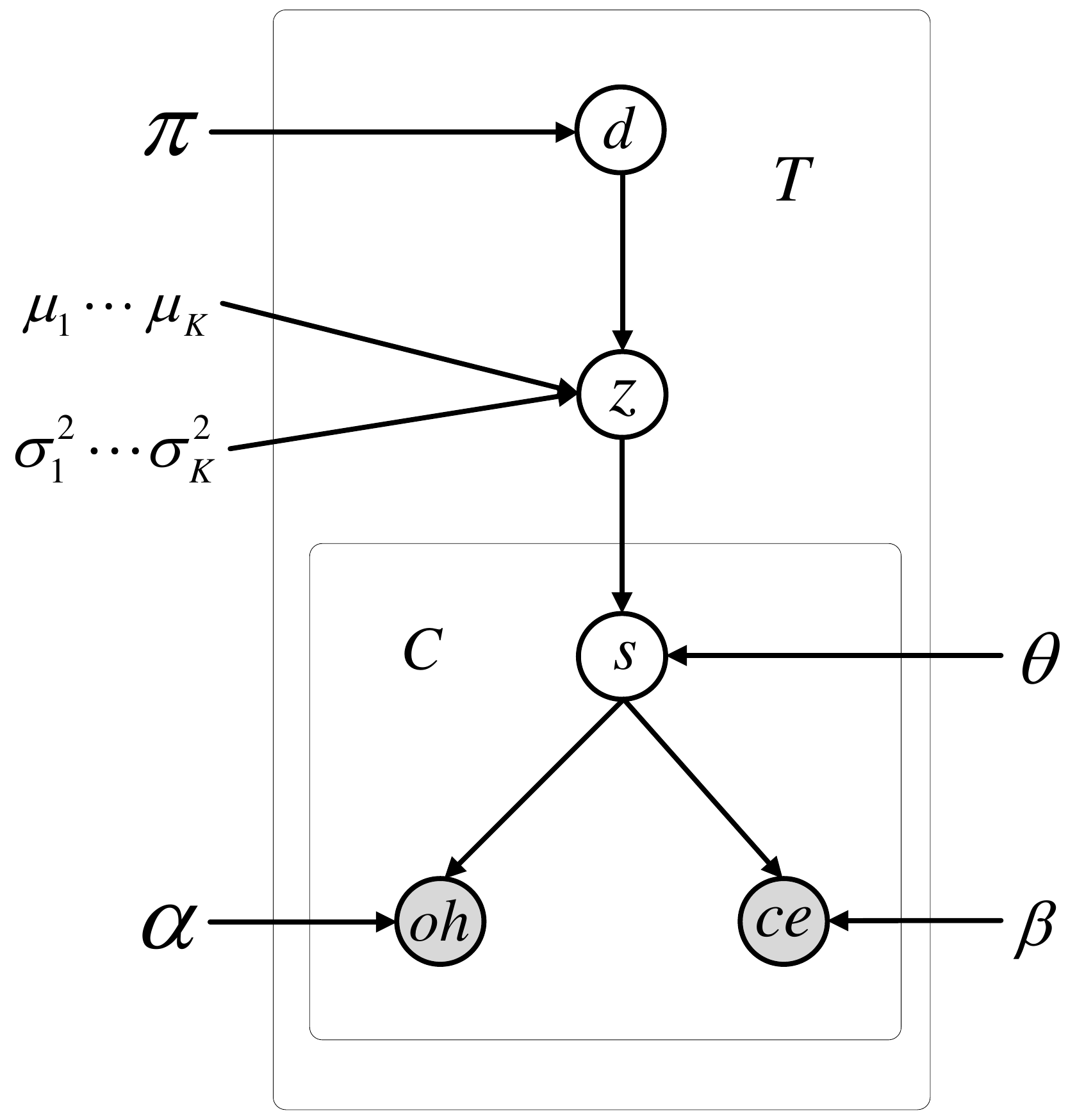}
        \label{fig:model_gmvae}
    }
\caption{Illustration of DSI models.
($T$ -- \# of user turns; $C$ -- \# of candidates per turn. $d$ and ${\bf s}$ are discrete latent variables, ${\bf z}$ is a continuous latent variable. ${\bf oh}$ and ${\bf ce}$ are observed data.)}
\label{fig:models}
\end{figure}

\subsection{Model 1: DSI-base}
Our first model is shown in Figure~\ref{fig:model_vae}. The model is a generative model, which explains the occurrences of values (as represented by both a one-hot vector and a contextualized dense vector) from a structured slot frame, distributed according to a hidden vector variable ${\bf z}$. This framework follows the variational dense embedding method of \newcite{kingma2014vae}. In particular, ${\bf z}$ is treated as a Gaussian vector, with the mean and variance factors themselves being decided according to the observed values. As a result, its training follows a variational auto-encoder scheme. 

Given a corpus of user turns $\mathcal{T}$, for each turn $t \in \mathcal{T}$, there is a set of candidate values ${C_t}$. We first sample a latent state vector ${\bf z}$ from a global Gaussian distribution. Then a neural network $f_s({\bf z};\boldsymbol{\theta})$ takes ${\bf z}$ as input to encode slot distribution logits, sampling a discrete slot assignment vector ${\bf s}$ corresponding to each candidate $c \in C_t$ in turn $t$. Last, for each candidate, we sample a one-hot vector ${\bf oh}$ from a categorical distribution parameterized by the output of a neural network $f_{oh}({\bf s};\boldsymbol{\alpha})$, as well as a contextualized word embedding vector ${\bf ce}$ from a multivariate Gaussian distribution parameterized by the output of a neural network $f_{ce}({\bf s};\boldsymbol{\beta})$. 

In particular, for the latent state ${\bf z}$, the Gaussian distribution is parameterized by a mean vector $\boldsymbol{\mu}$ and a diagonal covariance matrix $\boldsymbol{\sigma}^2$. Suppose that there are $S$ slots. The categorical slot distribution for each candidate $c$ is parameterized by a probability vector $\boldsymbol{\gamma} \in \mathbb{R}^{S}$. For each slot $s$, the Gaussian distribution for the contextualized embedding ${\bf ce}$ is parameterized by a mean vector $\boldsymbol{\mu}_s \in \mathbb{R}^{n}$ and a diagonal vector of covariance matrix $\boldsymbol{\sigma}_s \in \mathbb{R}^{n}$, where $n$ represents the dimension of ${\bf ce}$. Further, the categorical distribution for the one-hot vector ${\bf oh}$ is parameterized by a probability vector $\boldsymbol{\lambda}_s \in \mathbb{R}^{V}$, where $V$ is the candidate value vocabulary size. All the parameters are obtained through neural networks.


The generative process is shown in Algorithm~\ref{alg:vae}. Accordingly, the joint probability for a turn $t$ can be factorized as:
\begin{equation}
    p(t) = p({\bf z})\prod_{c \in C_t}p({\bf s}|{\bf z})p({\bf oh}|{\bf s})p({\bf ce}|{\bf s})
    \label{eqn:fact_t}
\end{equation}
where the probability terms are defined as:
\begin{eqnarray}
p({\bf z}) &=& \mathcal{N}\left({\bf z}|\boldsymbol{\mu},\boldsymbol{\sigma}^2\right)\label{eqn:z}\\
p({\bf s}|{\bf z}) &=& \textup{Cat}({\bf s}|\boldsymbol{\gamma})\label{eqn:s_z}\\
p({\bf oh}|{\bf s}) &=& \textup{Cat}({\bf oh}|\boldsymbol{\lambda}_s)\label{eqn:h_s}\\
p({\bf ce}|{\bf s}) &=& \mathcal{N}\left({\bf ce}|\boldsymbol{\mu}_s,\boldsymbol{\sigma}_s^2\right)\label{eqn:f_s}
\end{eqnarray}

$\boldsymbol{\mu}$ and $\boldsymbol{\sigma}^2$ are the parameters of a Gaussian prior distribution. $\boldsymbol{\gamma}$, $\boldsymbol{\lambda}_s$, $\boldsymbol{\mu}_s$ and $\boldsymbol{\sigma}_s^2$ are calculated as:
\begin{eqnarray}
\boldsymbol{\gamma} &=& \text{softmax}(W_{\gamma}{\bf z}+b_{\gamma})\\
\boldsymbol{\lambda}_s &=& \text{softmax}(W_{\lambda}{\bf s}+b_{\lambda})\\
\boldsymbol{\mu}_s &=& \text{BN}(W_{{\mu}}{\bf s}+b_{{\mu}})\\
\boldsymbol{\sigma}_s &=& \text{BN}(W_{{\sigma}}{\bf s}+b_{{\sigma}})
\end{eqnarray}


\begin{algorithm}[t]
\setlength{\belowcaptionskip}{-0.3cm}
\caption{\textit{DSI-base}}
\label{alg:vae}
\algsetup{indent=1em,linenosize=\footnotesize}
\footnotesize
\begin{algorithmic}[1]
    \FOR{each user turn $t \in \mathcal{T}$}
    	\STATE Sample a latent state vector ${\bf z} \sim \mathcal{N}\left(\boldsymbol{\mu},\boldsymbol{\sigma}^2\right)$
    	\FOR{each candidate $c \in C_t$}
	    	\STATE Compute a probability vector $\boldsymbol{\gamma} = f_{s}({\bf z};\boldsymbol{\theta})$
    	    \STATE Sample a slot vector ${\bf s} \sim \textup{Cat}(\boldsymbol{\gamma})$
    	    \STATE Compute a probability vector $\boldsymbol{\lambda}_s = f_{oh}({\bf s};\boldsymbol{\alpha})$
    		\STATE Sample a one-hot vector ${\bf oh} \sim \textup{Cat}(\boldsymbol{\lambda}_s)$
    	    \STATE Compute $[\boldsymbol{\mu}_s;\log \boldsymbol{\sigma}_s^2] = f_{ce}({\bf s};\boldsymbol{\beta})$
    		\STATE Sample a contextualized word embedding vector \\
    		${\bf ce} \sim \mathcal{N}\left(\boldsymbol{\mu}_s,\boldsymbol{\sigma}_s^2\right)$
    	\ENDFOR
    \ENDFOR
\end{algorithmic}
\end{algorithm}

\begin{algorithm}[t]
\setlength{\belowcaptionskip}{-0.3cm}
\caption{\textit{DSI-GM}}
\label{alg:gmvae}
\algsetup{indent=1em,linenosize=\footnotesize}
\footnotesize
\begin{algorithmic}[1]
    \FOR{each user turn $t \in \mathcal{T}$ }
        \STATE Sample a domain $d \sim \textup{Cat}(\boldsymbol{\pi})$
    	\STATE Sample a latent state vector ${\bf z} \sim \mathcal{N}\left(\boldsymbol{\mu}_d,\boldsymbol{\sigma}_d^2\right)$
    	\FOR{each candidate $c \in C_t$}
	    	\STATE Compute a probability vector $\boldsymbol{\gamma} = f_{s}({\bf z};\boldsymbol{\theta})$
			\STATE Sample a slot vector ${\bf s} \sim \textup{Cat}(\boldsymbol{\gamma})$
    	    \STATE Compute a probability vector $\boldsymbol{\lambda}_s = f_{oh}({\bf s};\boldsymbol{\alpha})$
    		\STATE Sample a one-hot vector ${\bf oh} \sim \textup{Cat}(\boldsymbol{\lambda}_s)$
    	    \STATE Compute $[\boldsymbol{\mu}_s;\log \boldsymbol{\sigma}_s^2] = f_{ce}({\bf s};\boldsymbol{\beta})$
    		\STATE Sample a contextualized word embedding vector \\
    		${\bf ce} \sim \mathcal{N}\left(\boldsymbol{\mu}_s,\boldsymbol{\sigma}_s^2\right)$
    	\ENDFOR
    \ENDFOR
\end{algorithmic}
\end{algorithm}

\subsection{Model 2: DSI-GM}
A limitation of \textit{DSI-base} is that sampling a latent state vector ${\bf z}$ from a global Gaussian distribution does not sufficiently model the fact that different domains may have different distributions. For those slots that appear in different domains (for example, slot {\it name} appear both in domain {\it restaurant} and {\it hotel} with similar utterance contexts), it can be difficult for {\it DSI-base} to distinguish them correctly. We apply a Gaussian Mixture Model (GMM) in \textit{DSI-base} by assuming the state vector is generated from a {\it Mixture-of-Gaussians}, in which each Gaussian represents a domain. This is inspired by Variational Deep Embedding (VaDE)~\cite{JiangZTTZ17}, which combines VAE and GMM for a clustering task.  

Specifically, suppose that there are $K$ domains. As shown in Figure~\ref{fig:model_gmvae}, we first sample a domain $d$ from a categorical distribution parameterized by $\boldsymbol{\pi}\in \mathbb{R}^K$, where $\pi_d$ is the prior probability for domain $d$ and $\sum_{d=1}^K \pi_d=1$. The latent state vector ${\bf z}$ is sampled from a Gaussian distribution with parameters of a mean vector $\boldsymbol{\mu}_d$ and a covariance vector $\boldsymbol{\sigma}_d$ corresponding to the chosen domain $d$. The latent state vector is then used for sampling a slot vector for each candidate in a turn, in the same way as \textit{DSI-base}.

According to the generative process shown in Algorithm \ref{alg:gmvae}, the joint probability for a user turn $t$ is
\begin{equation}
    p(t) = p(d)p({\bf z}|d)\prod_{c \in C_t}p({\bf s}|{\bf z})p({\bf oh}|{\bf s})p({\bf ce}|{\bf s}),
    \label{eqn:fact_t_d}
\end{equation}
where the additional probabilities are defined as:
\begin{eqnarray}
p(d) &=& \textup{Cat}(d|{\boldsymbol{\pi}})\label{eqn:d}\\
p({\bf z}|d) &=& \mathcal{N}\left({\bf z}|\boldsymbol{\mu}_d,\boldsymbol{\sigma}_d^2\right)\label{eqn:z_d}
\end{eqnarray}


\begin{figure}[!t]
	\setlength{\belowcaptionskip}{-0.3cm}
	\centering
	\includegraphics[width=0.48\textwidth]{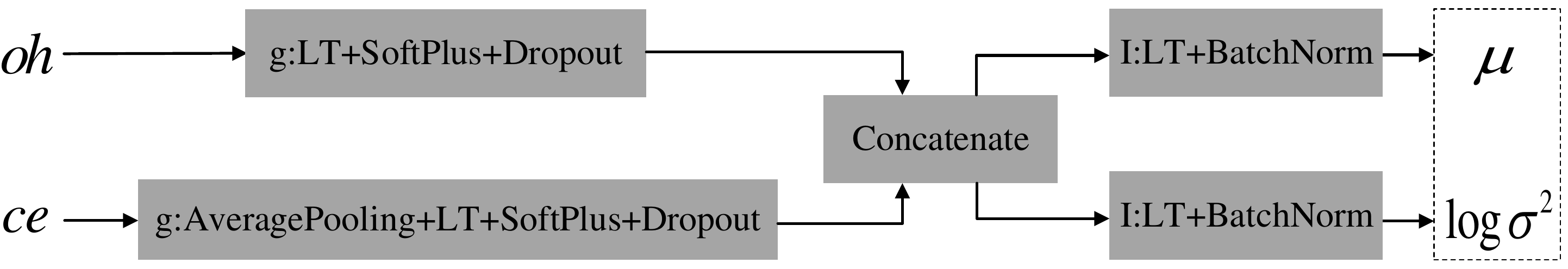}
	\caption{An encoder network is used to maximize the ELBO of {\it DSI-base} and {\it DSI-GM}. ``LT'' denotes linear transformation.}
	\label{fig:encoder}
\end{figure}

\subsection{Inference}
Given the generative process of {\it DSI-GM}, following~\newcite{liu-etal-2019-open}, we collapse the discrete slot latent variable ${\bf s}$ and rewrite the joint log-likelihood as:
\begin{flalign}
\log p(t)&=\log\int_{\bf z}\sum_{d}p(d)p({\bf z}|d)\prod_{c \in C_t} p({\bf oh}|{\bf z})p({\bf ce}|{\bf z})d{\bf z}\nonumber\\
&\geq E_{q({\bf z},d|{t})}[\log\frac{p(t,{\bf z},d)}{q({\bf z},d|{ t})}]=\mathcal{L}_{\textup{ELBO}}({t})\label{eqn:loglikelihood}
\end{flalign}
where $\mathcal{L}_{\textup{ELBO}}$ is the evidence lower bound (ELBO). Since direct optimization and inference for Equation~\ref{eqn:loglikelihood} is intractable. We follow previous work~\cite{kingma2014vae,JiangZTTZ17} and use a variational posterior distribution $q({\bf z},d|{ t})$ to approximate the true posterior distribution $p({\bf z},d|{ t})$. By assuming a mean field distribution~\cite{xing2002generalized}, $q({\bf z},d|{ t})$ can be factorized as $q({\bf z}|t)q(d|t)$. The posterior $q({\bf z}|t)$ can be modeled using a multivariate Gaussian distribution, with the mean vector $\boldsymbol{\mu}_d$ and the variance vector $\boldsymbol{\sigma}_d$ obtained through a neural network as shown in Figure~\ref{fig:encoder}. $q(d|t)$ is calculated as follows:
\begin{equation}
q(d|{t})=p(d|{\bf z})\equiv\frac{p(d)p({\bf z}|d)}{\sum_{d'=1}^Kp(d')p({\bf z}|d')}
\label{eqn:p_d_z}
\end{equation}

Using the reparameterization trick~\cite{kingma2014vae}, the ELBO can be decomposed into a {\it reconstruction} term and a {\it regularization} term, respectively:
\begin{equation}
\mathcal{L}_{\textup{ELBO}}({t})=E_{q({\bf z},d|{t})}[\log p(t|{\bf z})]-D_{KL}(q({\bf z},d|{t})||p({\bf z},d))
\label{eqn:elbo}
\end{equation}
where $D_{KL}(q({\bf z},d|{t})||p({\bf z},d))$ is the KL divergence between the \textit{Mixture-of-Gaussians} prior $p({\bf z},d)$ and variational posterior $q({\bf z},d|{t})$.

After training, the domain and slot assignment for each candidate $c$ can be obtained by Equation~\ref{eqn:p_d_z} and as follows:
\begin{align}
p({\bf s}|c,t)=&p({\bf s}|c,{\bf z}) \propto p({\bf s},{\bf oh},{\bf ce},{\bf z}) \\
=& \,p({\bf s}|{\bf z})p({\bf oh}|{\bf s})p({\bf ce}|{\bf s}) 
\label{eqn:p_s_z}
\end{align}

\section{DSI-Based Response Generation}
\begin{figure}[!t]
\setlength{\belowcaptionskip}{-0.2cm}
\centering
\includegraphics[width=0.48\textwidth]{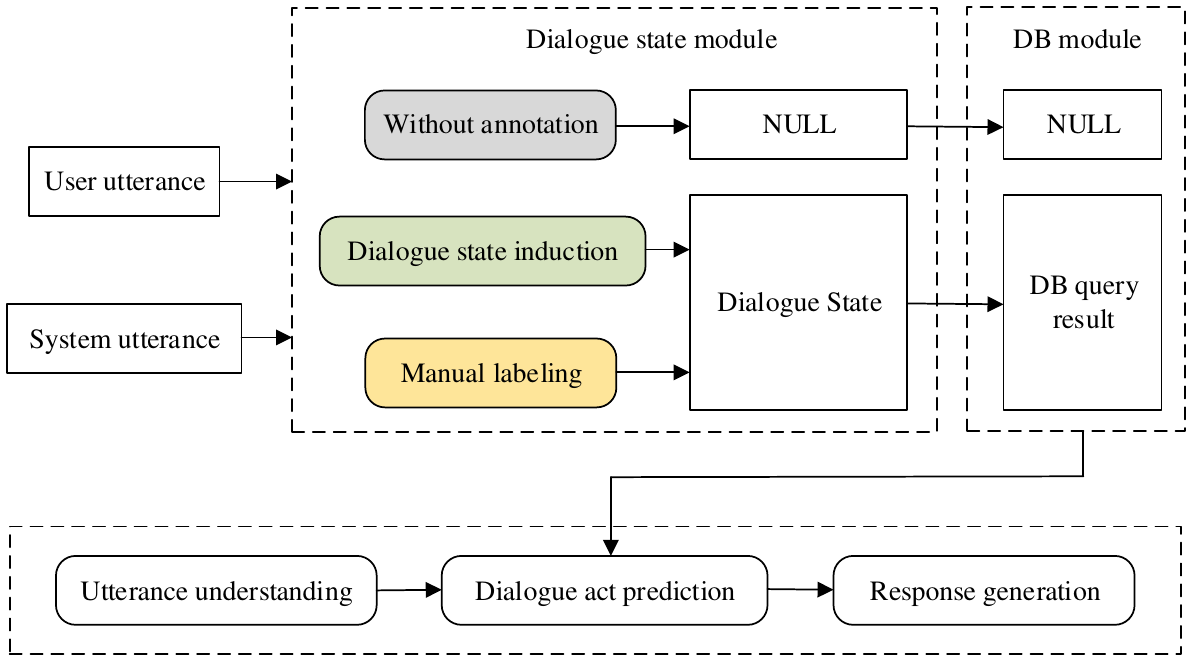}
\caption{DSI-based dialogue response generation.}
\label{fig:dsi-hdsa}
\end{figure}

We apply the DSI models on the task of downstream response generation, following the pipeline system which was first proposed by~\newcite{wen2017network} and then decomposed into two components by \newcite{chen-etal-2019-semantically}. As shown in Figure~\ref{fig:dsi-hdsa}, the top component consists of the dialogue state module and database operation module, while the bottom component contains the dialogue act prediction module and response generation module. For the bottom component, we directly take the recently proposed HDSA model~\cite{chen-etal-2019-semantically} to test our induced dialogue states. 


For the top component, we investigate three different settings of the dialogue state module: (\romannumeral1) empty dialogue states under the condition that no annotation is available (grey oval), (\romannumeral2) dialogue states induced by our mixture model (green oval), (\romannumeral3) dialogue states obtained by manual labeling (golden oval). In the last two settings, the dialogue states are regarded as input for subsequent modules, while in the first setting, no state information is given, which corresponds to end-to-end models~\cite{eric2017copy,wu2019globaltolocal} for dialogue system. The first setting is used to test the effectiveness of our model, while the third setting can be regarded as the oracle upper bound of our model. 

\section{Experiments}

We evaluate our proposed DSI task and its effectiveness on the downstream tasks using the MultiWOZ 2.1~\cite{eric2019multiwoz} dataset, which fixes some noisy state annotations in the MultiWOZ 2.0~\cite{budzianowski-etal-2018-multiwoz} dataset.
MultiWOZ2.1 contains 10,438 multi-turn dialogues and we follow the same partition as \newcite{WuTradeDST2019}. To justify the generalization of the proposed model, we also use a recently proposed SGD~\cite{rastogi2019towards} dataset, which contains 16,142 multi-turn dialogues and is the largest existing conversational corpus. We use the same train/validation split sets as~\newcite{rastogi2019towards}. 
%

\begin{table*}[t]
\setlength{\belowcaptionskip}{-0.3cm}
\centering
\small
\resizebox{\textwidth}{!}{%
\begin{tabular}{lcccccccc|cccc|cccc}
\toprule
\multirow{3}{*}{\textbf{Models}} & \multicolumn{8}{c|}{\textbf{MultiWOZ 2.1}} & \multicolumn{8}{c}{\textbf{SGD}} \\ \cmidrule{2-17} 
 & \multicolumn{4}{c|}{\textbf{Turn level}} & \multicolumn{4}{c|}{\textbf{Joint level}} & \multicolumn{4}{c|}{\textbf{Turn level}} & \multicolumn{4}{c}{\textbf{Joint level}} \\ \cmidrule{2-17} 
 & \textbf{Precision} & \textbf{Recall} & \textbf{F1} & \multicolumn{1}{c|}{\textbf{Accuracy}} & \textbf{Precision} & \textbf{Recall} & \textbf{F1} & \textbf{Accuracy} & \textbf{Precision} & \textbf{Recall} & \textbf{F1} & \textbf{Accuracy} & \textbf{Precision} & \textbf{Recall} & \textbf{F1} & \textbf{Accuracy} \\ \midrule
{\it Random} & 1.49 & 1.51 & 1.49 & \multicolumn{1}{c|}{1.39} & 0.21 & 0.28 & 0.23 & 0.02 & 0.94 & 0.95 & 0.94 & 0.92 & 0.05 & 0.08 & 0.06 & 0.02 \\
{\it DSI-base} & 38.8 & 37.7 & 37.3 & \multicolumn{1}{c|}{25.7} & 33.9 & 32.1 & 32.1 & 2.3 & 27.0 & 26.0 & 26.0 & 21.1 & 13.9 & 17.5 & 14.5 & 2.3 \\
{\it DSI-GM} & 52.5 & 39.3 & 49.6 & \multicolumn{1}{c|}{36.1} & 49.2 & 43.2 & 44.8 & 5.0 & 34.7 & 33.4 & 33.5 & 27.5 & 19.0 & 22.9 & 19.5 & 3.1 \\ 
\bottomrule
\end{tabular}%
}
\caption{Overall results of DSI.}
\label{tab:overall}
\end{table*}

\begin{table}[t]
\setlength{\belowcaptionskip}{-0.3cm}
\centering
\resizebox{0.45\textwidth}{!}{%
\begin{tabular}{c|c|l|c|c}
\toprule
{\bf Name} & \multicolumn{2}{c|}{{\bf Value}} & {\bf Name} & {\bf Value} \\ 
\midrule
Domain number ({\it DSI-GM} only) & \multicolumn{2}{c|}{100} & Batch size & 200 \\
Slot number ({\it DSI-base}/{\it DSI-GM}) & \multicolumn{2}{c|}{300/1000} & Dropout & 0.2 \\
Feature dimension & \multicolumn{2}{c|}{256} & Learning rate & 0.02 \\
Linear transformation layer size & \multicolumn{2}{c|}{100} & Momentum & 0.99 \\ 
\bottomrule
\end{tabular}%
}
\caption{Hyper-parameters settings.}
\label{tab:param}
\end{table}

\subsection{Experimental Settings}
For the DSI models, the Adam optimizer is used to maximize the ELBO of Equation~\ref{eqn:elbo}. All the models are trained over the training set, where hyper-parameters are tuned on the development set, before being finally used on the test set. Since no manual labels are available, we follow~\newcite{liu-etal-2019-open} and select the hyper-parameters which fit the best ELBO score on the dev set as shown in Table~\ref{tab:param}. 
For the downstream HDSA model, we directly take the original hyper-parameters. Since our datasets are manually labeled with domains and slots ({\it restaurant-name}), we can name the induced slots after the gold-standard slots that have the maximum value match. In addition, in the datasets, each slot is labeled with a service domain, and thus we obtain a domain output also.


\subsection{Evaluation Metrics}
\paragraph{DSI} For each turn, {\it DSI-base} and {\it DSI-GM} induce several or no slot-value pairs based on the current user utterance and its preceding system utterance. We compare the DSI outputs with the slots that have non-empty assignments in the ground truth dialogue states for the current user turn. We consider the following two metrics\footnote{A fuzzy matching mechanism is used to compare induced values with the ground truth.}:
\begin{enumerate}
    \item \textbf{State Matching} (\textit{Precision}, \textit{Recall} and \textit{F1-score} in Table~\ref{tab:overall}): Similar to previous work ~\cite{liu-etal-2019-open}, we use state matching to evaluate the overlapping of induced states and the ground truth.
    \item \textbf{Goal Accuracy} (\textit{Accuracy} in Table~\ref{tab:overall}): We adopt this standard metric from DST~\cite{WuTradeDST2019,zhong2018global}. The predicted dialogue states for a turn is considered as true only when all the user search goal constraints are correctly and exactly identified.  
\end{enumerate} 

We evaluate both metrics in both the {\bf turn level} and the {\bf joint level} (Table~\ref{tab:overall}). The joint level metrics are more strict in jointly considering the output of all turns.

\paragraph{Response generation} Following \newcite{chen-etal-2019-semantically}, the dialogue act prediction results are evaluated in terms of {\it Precision}, {\it Recall} and {\it F1-score}. Delexicalized-BLEU and Entity F1 are used to evaluate response generation.

\subsection{Results}
\paragraph{DSI Performance} The DSI results are shown in Table \ref{tab:overall}. We have four main observations:
	\begin{itemize}
	\item Both {\it DSI}  models show great advantages over a random select strategy, which randomly assigns a reference slot for each candidate. This shows the strength of our neural generative models with hidden variables.
	\item \textit{DSI-GM} outperforms \textit{DSI-base} on both the turn level and joint level metrics, which demonstrates the effectiveness of the GMM model, which finds an appropriate domain first before sampling a latent state representation. We attribute this to the fact that the dialogue states can be more effectively regarded as a hierarchical structure (i.e., domain-slot-value) and hence first detecting a domain and then a slot under this domain can help alleviate the difficulty of distinguishing the appropriate slot in a large mixture of similar slot values.
	\item The {\it joint goal accuracy} is significantly lower compared with the other metrics, which shows that the metric can be overly strict in our \textit{unsupervised} setting. This is a consistent observation of recent work on cross-lingual dialogue state tracking~\cite{liu2019attentioninformed}, which shows that the joint goal accuracy of a cross-lingual DST model can be as low as 11\% on accuracy even with cross-lingual contextualized embeddings. Furthermore, the \textit{joint goal F1-score} can reach 44.8\% on MultiWOZ dataset, which shows that our model can achieve promising performance without any labeled training data.
	\item The results among all metrics on the SGD dataset are lower than those on the MultiWOZ2.1 dataset. We attribute it to this reason that the SGD dataset is more difficult than the MultiWOZ dataset because it contains more types of domains and slots (16 domains and 214 slots as compared with 7 domains and 24 slots in MultiWOZ).
	\end{itemize}


\begin{table}[t]
	\setlength{\belowcaptionskip}{-0.2cm}
	\small
	\centering
	\resizebox{0.45\textwidth}{!}{%
		\begin{tabular}{lccc|cc} 
			\toprule
			\multirow{2}{*}{\bf Dialogue State} & \multicolumn{3}{c|}{\bf Dialog Act Prediction} & \multicolumn{2}{c}{\bf Delexicalized}\\ \cmidrule{2-6} 
			& {\bf Precision} & {\bf Recall} & {\bf F1}  & {\bf BLEU} & {\bf Entity F1}\\ 
			\midrule
			{\it None} & 71.0 & 67.4 & 69.1  & 18.7 & 54.6\\
			{\it DSI-GM} & 72.0 & 70.5  & 71.2 & 20.8 & 56.5\\
			{\it Manual labeling} & 75.6 & 73.0 & 74.2  & 21.6 & 61.3\\
			\bottomrule
		\end{tabular}
	}
	\caption{Empirical results on MultiWOZ dialogue act prediction and response generation.}
	\label{tab:generate}
\end{table}

\paragraph{DAP and Response Generation} The dialogue act prediction (DAP) and response generation results are shown in Table~\ref{tab:generate}.  Compared to not using dialogue states, the outputs of {\it DSI-GM} allows a subsequent model~\cite{chen-etal-2019-semantically} to improve the F1-score of dialogue act prediction by 2.1\%, and further improve the BLEU and entity F1 for a system output utterance by 2.1\% and 1.9\%, respectively. This shows that maintaining dialogue states can be useful in neural dialogue systems, as consistent with observations from DST research~\cite{lei-etal-2018-sequicity,wen-etal-2018-sequence}. The results demonstrate that DSI is a useful task in dialogue systems research and our baseline models are effective. In Table~\ref{tab:generate}, the gap between {\it DSI-GM} and {\it manual labeling} indicates further rooms for improvement on the dialogue model.

\begin{figure}[!t]
	\centering
	\includegraphics[width=0.40\textwidth]{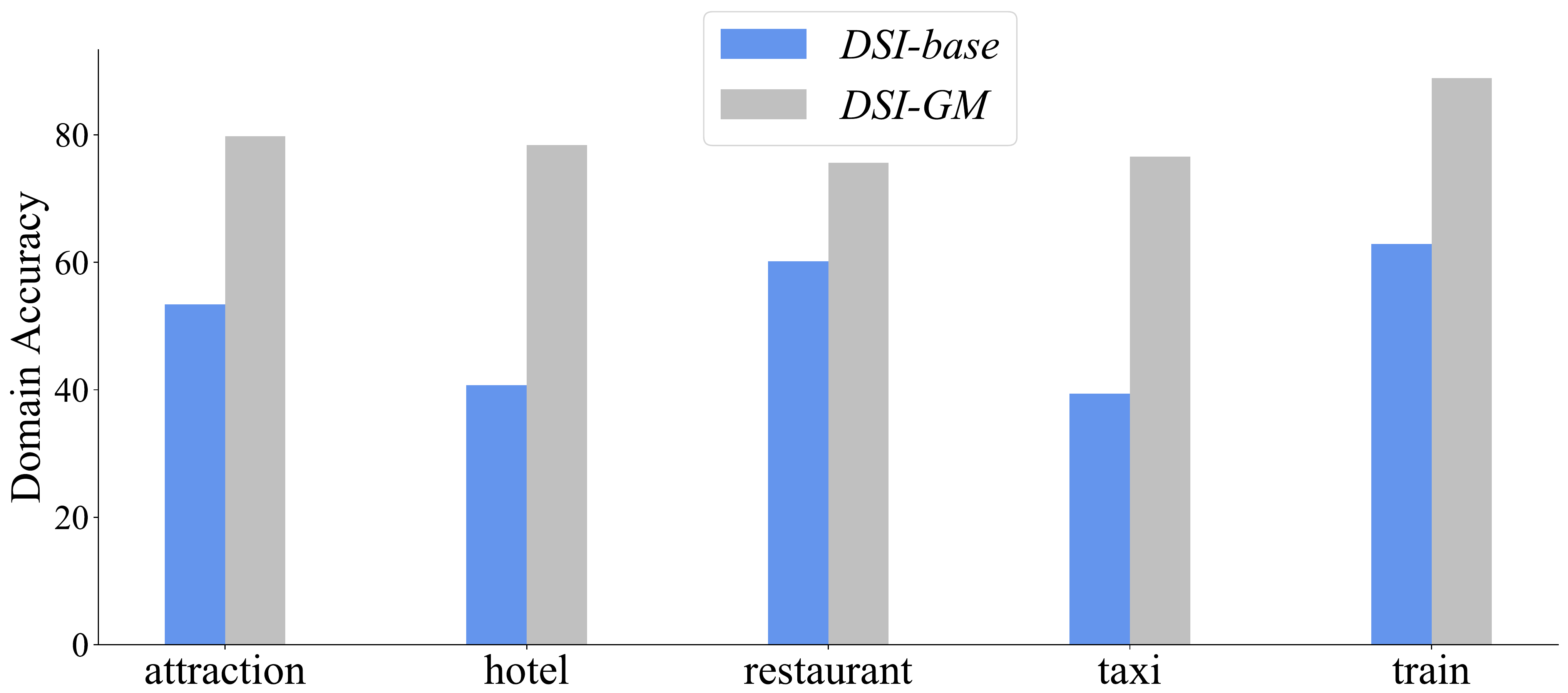}
	\caption{Domain accuracy.}
	\label{fig:domain_accu}
\end{figure}
\begin{figure}[t]
    \centering
    \subfigure[\textit{DSI-base}]{
        \centering
        \includegraphics[width=0.22\textwidth]{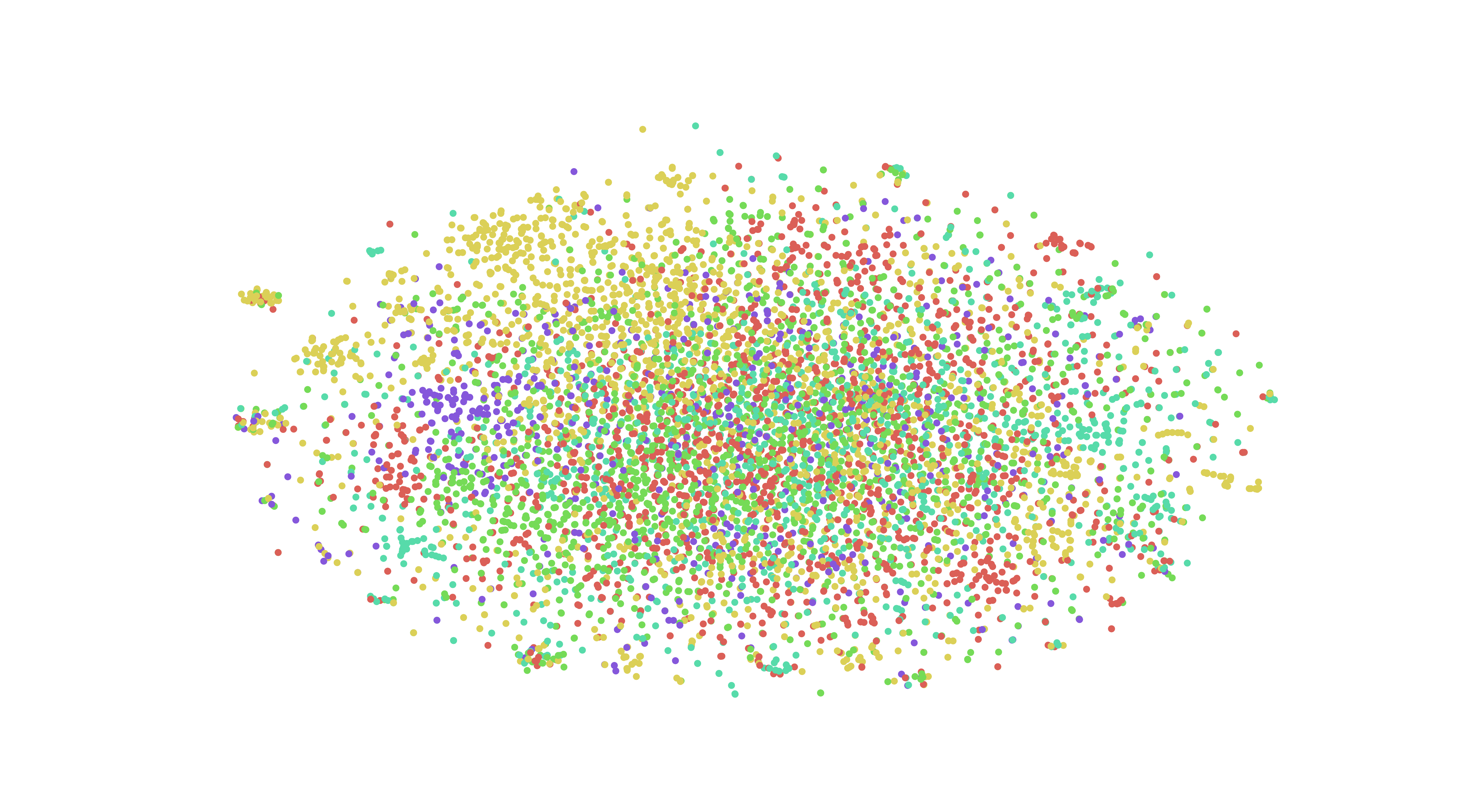}
        \label{fig:visu_vae}
    }
    \subfigure[\textit{DSI-GM}]{
        \centering
        \includegraphics[width=0.22\textwidth]{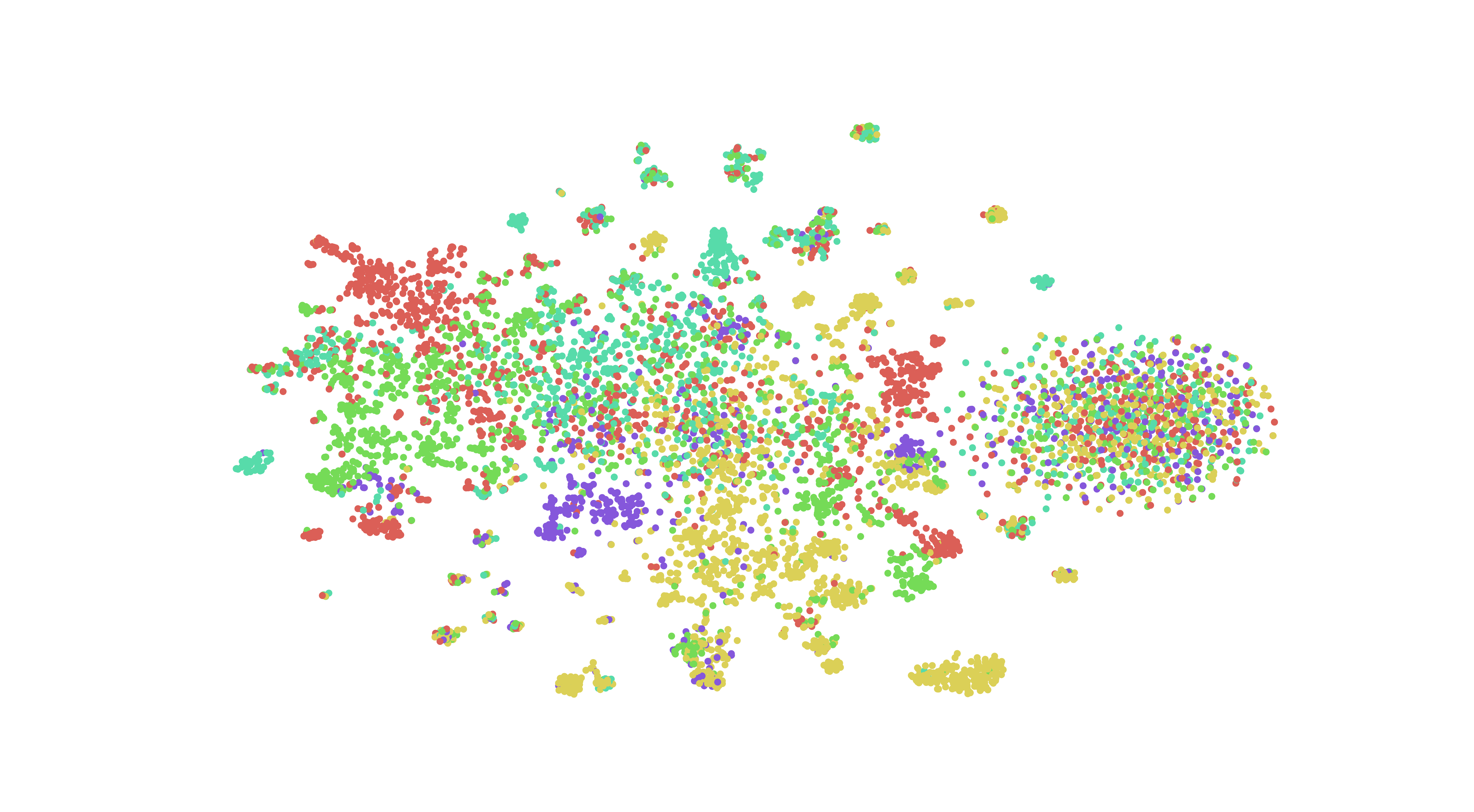}
        \label{fig:visu_gmvae}
    }
\caption{Visualization of state vectors learned by {\it DSI-base} and {\it DSI-GM}. Each color represents a domain.}
\label{fig:visu}
\end{figure}

\begin{figure}[!t]
	\centering
	\includegraphics[width=0.47\textwidth]{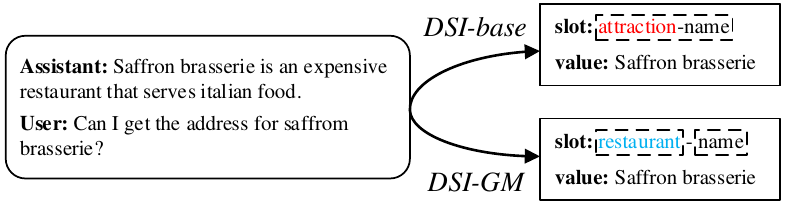}
	\caption{Example output by {\it DSI-base} and {\it DSI-GM}. The {\color{cyan}cyan domain} is correct, while the {\color{red}red domain} is wrong.}
	\label{fig:case}
\end{figure}

\subsection{{\it DSI-GM} vs. {\it DSI-base}}
\paragraph{Domain Accuracy} We measure the fraction of user turns for which the domains are correctly induced for all candidates. Figure~\ref{fig:domain_accu} shows the comparison between the {\it DSI-base} model and the {\it DSI-GM} model. We can see that the \textit{DSI-GM} outperforms the \textit{DSI-base} on each domain, which demonstrates that explicitly modeling the domain distribution is effective across all domains. In addition, to better intuitively know how much better the domain representation can be modeled through \textit{DSI-GM}, we visualize the learned state representations of \textit{DSI-base} and \textit{DSI-GM} only with its domain label. In particular, we use t-SNE to reduce the dimensionality of the latent representation ${\bf z}$ and plot the whole test set in Figure~\ref{fig:visu}. Compared with \textit{DSI-base}, whose latent state representations are mixed in the domain level, \textit{DSI-GM} shows its superiority in representing different domains, as representations are clustering in a more orderly way. This further demonstrates the effectiveness of \textit{GMM} applied in our task.

\paragraph{Case Study} Figure~\ref{fig:case} shows a case on the dialogue states induced by {\it DSI-base} and {\it DSI-GM}. In this case, both the \textit{attraction} and \textit{restaurant} domains consist of a \textit{name} slot, which shares similar contexts such as ``Can i get the address for''. This can make their contextualized features similar although the two {\it name} slots should be treated as two distinct types.
{\it DSI-base} generates an incorrect domain \textit{attraction} while {\it DSI-GM} induces the correct domain {\it restaurant}. We attribute it to the fact that the {\it DSI-GM} model captures domain information by explicitly modeling the multi-domain distribution through a \textit{Mixture-of-Gaussians} instead of a global \textit{Gaussian}.

\subsection{Error Analysis}
\begin{figure}[!t]
	\centering
	\includegraphics[width=0.45\textwidth]{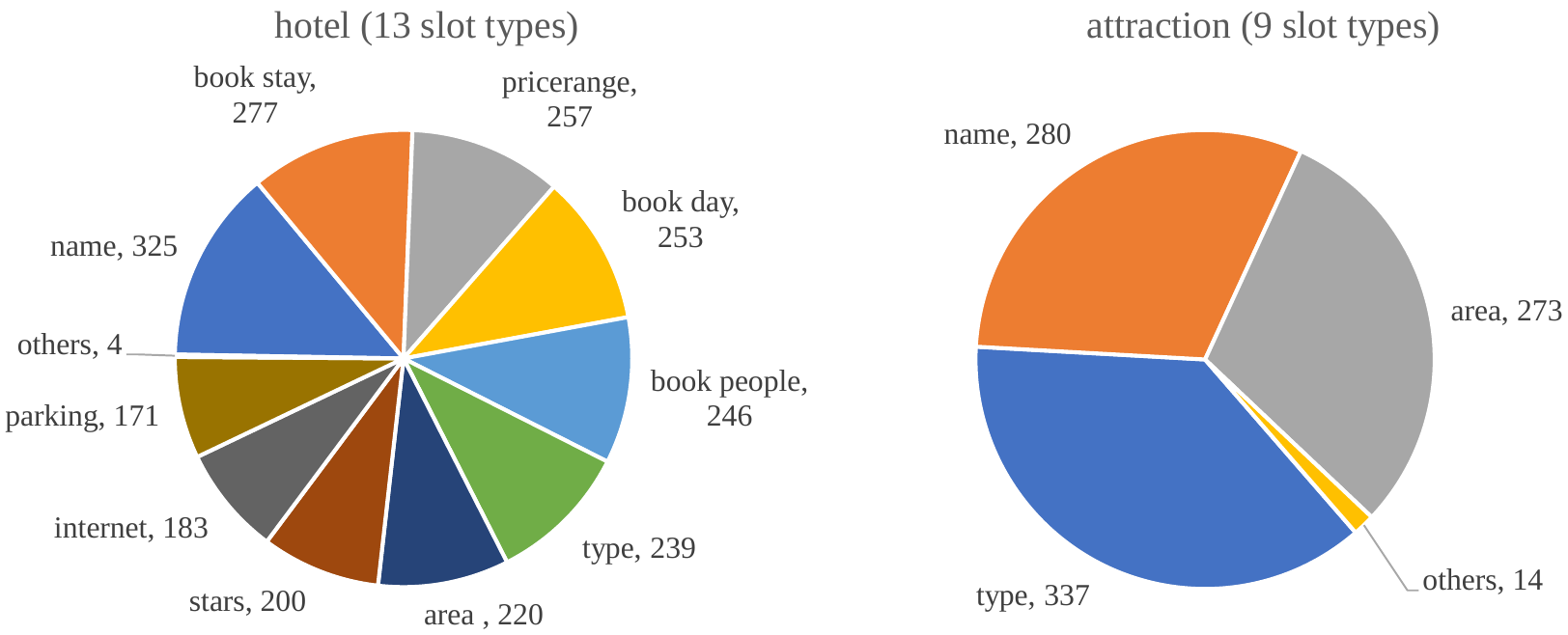}
	\caption{Slots distributions on {\it hotel} and {\it attraction} domains.}
	\label{fig:slot_distri}
\end{figure}

We present the accuracies on each domain between \textit{DSI-base} and \textit{DSI-GM}. The results are shown in Table~\ref{tab:error_analysis}. Both \textit{DSI-base} and \textit{DSI-GM} give the lowest accuracy on the {\it hotel} domain and a relatively higher accuracy on the {\it attraction} domain. To further understand the reason, we calculate the statistics of slots on the {\it hotel} and {\it attraction} domains, which are shown in Figure~\ref{fig:slot_distri}. It can be seen that the numbers of dominant slot types of the two domains are 10 and 3, respectively, correctly recognizing which can give strong overall accuracies. In both domains, these slots are distributed evenly. This indicates that the number of distinct slots is a key factor to the difficulty level for our DSI models, which is intuitive.

\begin{table}[!t]
\centering
\resizebox{0.40\textwidth}{!}{%
\begin{tabular}{lccccc}
\toprule
 & attraction & hotel & restaurant & taxi & train \\
\midrule
{\it DSI-base} & 27.9 & 21.7 & 26.1 & 30.7 & 26.0 \\
{\it DSI-GM} & 40.3 & 31.4 & 35.6 & 39.9 & 36.8 \\
\bottomrule
\end{tabular}%
}
\caption{Turn goal accuracy per domain.}
\label{tab:error_analysis}
\end{table} 




\section{Conclusion}
We proposed a novel task of \textit{dialogue state induction}, which is to automatically identify dialogue state slots and values over a large set of dialogue records. Compared with existing research, our task is practically more useful for handling the large variety of services available in the industry, which disallows scalable manual labeling of dialogue states. We further built two neural generative models with latent variables. Results on standard DST datasets show that the models can effectively induce meaningful dialogue states from raw dialogue data, and further improve the results of a dialogue system compared to without using dialogue states. Our methods can serve as baselines for further research on the task. 

\section*{Acknowledgments}
We thank the anonymous reviewers for their detailed and constructive comments. The first three authors contributed equally. Yue Zhang is the corresponding author. 
We would like to acknowledge funding support from National Natural Science Foundation of China under Grant No.61976180 and the Westlake University and Bright Dream Joint Institute for Intelligent Robotics.

\appendix

\bibliographystyle{named}
\bibliography{ijcai20}

\end{document}